\documentclass[conference]{IEEEtran}
\IEEEoverridecommandlockouts
\usepackage{cite}
\usepackage{amsmath,amssymb}
\usepackage{booktabs}
\usepackage{array}
\usepackage{url}
\usepackage[hidelinks]{hyperref}
\usepackage{xcolor}

\begin{document}

\title{How Much of a Calculation-Error Rate Is Not\\
Arithmetic: Auditing the FinReason Cup 2026\\
Task 3 Metric with a Deterministic Rule System\\
\large FinReason Cup 2026, Task 3 Working Notes}

\author{\IEEEauthorblockN{Vladimir Beskorovainyi}
\IEEEauthorblockA{Besk Tech \\
besk.tech}}

\maketitle
\renewcommand{\thefootnote}{}
\footnotetext{These Working Notes describe team besk.tech's entry to Task~3 of the FinReason Cup 2026, held with IEEE Big Data 2026. They are posted at the organisers' request so that the task overview can cite them.}
\renewcommand{\thefootnote}{\arabic{footnote}}

\begin{abstract}
Task 3 of the FinReason Cup 2026 scores a system on two values per case, the one an XBRL filing reports and the one its own calculation relationships imply, and labels each failure as structural, extraction or calculation. This report shows that the calculation-error rate so defined does not count arithmetic failures, and bounds how far it is from one using the published board alone. One of the three rules behind the benchmark, DQC\_US\_0015, has an answer that is the negation of the reported value --- in all 110 published practice cases, against 1 of 120 and 6 of 102 for the other two --- so a calculation label on such a case records a wrong fact or a wrong route, not a wrong sum. Subtracting the available structural and extraction labels from each entry's error count on that rule gives a floor: the bound is positive for 22 of the 23 ranked entries, and at the top it is sharpest, with at least 11 of 12 and 12 of 13 calculation errors of the two joint-first entries falling on cases where no arithmetic remained. The calculation-error rate of the best systems on this task is almost entirely not arithmetic.

The vehicle for that result is an entry that reconstructs the three data-quality rules the benchmark is drawn from, DQC\_US\_0015, DQC\_US\_0117 and DQC\_US\_0126, as a deterministic system with no learned parameters, no training step and no external service, running all 680 cases in 1.4 seconds. It scores 96.62\% under the offline judge on development and 95.29\% on the published test set, and its residual error concentrates in cases where a concept carries several equally flagged facts and the question names no dimensional member. Reading those error classes against the per-rule accuracies is what exposed the metric, which is why both are reported here.
\end{abstract}


\section{Introduction}
Public companies file financial statements in XBRL, a format that binds each reported number to a taxonomy concept, a reporting period and, where relevant, dimensional members. XBRL~US, the United States jurisdiction of XBRL International, publishes the Data Quality Committee (DQC) rules, an open rule set that flags filings whose numbers violate expected accounting relationships, for example a cost reported as negative or a dimensional breakdown that disagrees with its total.

Task 3 of the FinReason Cup 2026, \emph{Financial Audit Verification}~\cite{task3overview}, turns this activity into a benchmark. Each case supplies one filing, rendered as its schema, its presentation, calculation, definition and label linkbases, its instance document and the relevant US GAAP taxonomy, followed by two questions about one concept and period:
\begin{itemize}
  \item \textbf{the reported value}: the figure as it appears in the instance document;
  \item \textbf{the calculated value}: the figure implied by the filing's calculation relationships.
\end{itemize}
A submission returns, per case, a two-key object \verb|{extracted_value, calculated_value}|.

The evaluation is hierarchical. The first failing check decides the label: \textbf{S} if the answer is not a valid two-key object, \textbf{E} if the reported value does not match the filing, \textbf{C} if the reported value is right but the calculated value is not, and \textbf{A} if both match. Values are compared by numeric meaning, so $-1{,}284$ and $-1284$ are equal, and the calculated value is held to zero tolerance~\cite{leaderboard}. Accuracy (ACC), the share of cases labelled \textbf{A}, is the headline metric; the structural, extraction and calculation error rates (SER, EER, CER) report where the remaining cases fail.

Two judges implement this scheme and both are used below, so it is worth separating them. The \emph{deterministic judge} ships with the starter kit and compares values by parsing them as numbers; it is available offline and is what a participant can run locally. The \emph{official judge} is a language model, \texttt{openai/gpt-5-mini} with prompt \texttt{finmr-judge-v1} under evaluation version 1.1~\cite{leaderboard}; it produced the leaderboard and the final standing. What its prompt receives beyond the two values and the reference answer I did not verify, and the reading in Section~\ref{sec:judges} depends on it: a judge that sees the filing might be accepting a defensible alternative fact rather than applying a looser standard. The two do not agree case for case. Section~\ref{sec:judges} shows that they disagree about error \emph{class} in a systematic and measurable direction, which matters because only one of them is available while the work is being done.

This report makes one claim about the task's metric and three about building for it. The claim is Section~\ref{sec:bound}: the official calculation-error rate is not a count of arithmetic failures, and the published board bounds how far it is from one, for every ranked entry and most sharply for the entries that won. It needs no access to the data, only the published rates and the definition of the rule they are reported against. It fell out of reading my own error classes against the per-rule accuracies, which is why the system that produced them is reported here in full.

The three points about building are these. Modelling the three underlying DQC rules directly, rather than reasoning case by case, produces a compact and accurate system. Its residual error concentrates in a class the shipped materials leave underdetermined, which neither the live DQC engine nor the general-purpose models I tried settled. And two negative results constrain how such a system can be developed at all: the released practice set is not a reliable proxy for the ranked phases, since a change that improved its score and agreed with the live rule engine cost one case of 680 on the development leaderboard; and the two judges classify errors differently enough that the offline breakdown cannot be used to decide what to work on.

\section{Task and Data}
\label{sec:data}
The two ranked phases each contain 680 cases~\cite{task3overview}. The development phase returns a score and a public ranking; the test phase returns only an acceptance receipt and decides the final result. A separate practice set of 332 cases is released with its reference answers, drawn from the FinMR corpus~\cite{finmr}; the ranked phases share no filing with it.

The practice filings are abbreviated, and this matters for how the set can be used. The instance document is cut off mid-document in more than half of its 332 cases, and in 132 of them the fact the question asks about is absent from the shipped text altogether, because the reference answer was computed on the complete filing. At most 200 of the 332 practice cases are therefore answerable from what is shipped, a ceiling of 60.24\%. The ranked phases do not carry this limit. In the revision used for both phases no instance document is cut off, and a fact matching the asked concept and period is present in all 680 cases of each, so the obstacle the practice set presents does not arise there.

The three DQC rules behind the benchmark appear in the practice set in the proportion 110 (DQC\_US\_0015), 120 (DQC\_US\_0117) and 102 (DQC\_US\_0126). The reported value for a DQC\_US\_0015 case is always negative in the practice set, 110 cases of 110, since the rule exists to flag a sign violation. The other two rules carry no such regularity: among the practice cases the DQC\_US\_0117 answers are 78 positive, 27 negative and 15 zero, and the DQC\_US\_0126 answers are 61 negative against 41 positive. Sign alone therefore identifies one rule and not the others.

The organisers publish their baselines on the task leaderboard~\cite{leaderboard}. On development under the deterministic judge, a sign-flip shortcut that negates whatever it extracts scores 19.85\%, and a rule-based baseline that sums the calculation children, or else falls back to the taxonomy balance, reaches 31.76\%. The rule-based baseline is also published under the official judge, where it scores 33.09\%; Section~\ref{sec:judges} makes use of that pair. These figures are the floor for the task.

\section{Related Work}
\label{sec:related}
This work builds directly on the rule set. The XBRL~US Data Quality Committee publishes its checks as open source in the \texttt{xule} rule language, together with the taxonomies of excluded members and axes each rule consults~\cite{dqc}, and Arelle executes them against a filing~\cite{arelle}. The reference answers here are therefore not human annotations but the output of an open program, so a system that reproduces the program is measured against a program rather than against an annotator.

Benchmarks such as FinQA~\cite{finqa}, TAT-QA~\cite{tatqa} and FinanceBench~\cite{financebench} treat a filing as text and tables to be read. Here the evidence is the structured instance with its calculation linkbase, assembled from the FinMR corpus~\cite{finmr}, and the hard part of extraction is not reading a figure but choosing among facts the structure makes equally valid. The ranking judge is a language model scoring a pair of numbers, and the reliability of such judges is an active question~\cite{llmjudge}; what this task offers that most settings do not is the same baseline scored by a language-model judge and by a deterministic one on the same cases, so the disagreement can be measured rather than estimated.

\section{Method}
The system is a single deterministic pipeline. It carries no learned parameters and no per-case branching. Most decisions follow from the filing and from the published definition of the three rules; where the rule text left a choice undetermined I added a tie-break, and Sections~\ref{sec:totals} and~\ref{sec:cand} say which. Two design choices were nonetheless selected against development-leaderboard feedback, with no held-out set to check them on; Section~\ref{sec:dev2test} reports what that appears to have cost.

\subsection{Parsing}
For each case I parse the instance document into facts, each with its concept, its context (period and dimensional members) and its numeric value, and I parse the calculation linkbase into weighted parent--child relationships per extended-link role. The asked-about concept and period are read from the question text.

\subsection{Rule DQC\_US\_0015: non-negative items}
\label{sec:nonneg}
DQC\_US\_0015, published as \emph{Negative Values}, lists concepts that should not be negative. When the asked concept is on that list, the flagged fact is by definition a negative one. I prefer a non-dimensional negative fact for the period; when none exists, I take a negative fact carried on a dimensional context, after removing members and axes that the rule excludes. The exclusion lists (members such as adjustment, consolidation and elimination members; axes such as the investment-identifier and equity-components axes) are reproduced from the rule's own taxonomy. The rule itself only flags, it does not prescribe a correction; the benchmark's reference answer for such a case is the negation of the reported one in every published instance, which Section~\ref{sec:bound} states precisely because an argument there depends on it.

\subsection{Rules DQC\_US\_0117 and DQC\_US\_0126: totals}
\label{sec:totals}
DQC\_US\_0117, published as \emph{Financial Statement Tables Dimensional Cross Check}, compares the members of an axis against the total they should sum to. DQC\_US\_0126, published as \emph{FS Calculation Check with Non Dimensional Data}, is a calculation check on non-dimensional data: it compares a concept's reported value against the sum its calculation children imply. What the two have in common, for this task, is that the asked value is usually the undimensioned total for the concept and period rather than any single member; the one exception is the second assertion of DQC\_US\_0117, noted below. The computations differ. For DQC\_US\_0126 the calculated value is the signed sum of the concept's calculation children for the asked period under the governing role. For DQC\_US\_0117 it is the sum of the facts the instance carries on the members of one axis for that concept and period; the definition linkbase enters only as a filter, supplying the members a role declares and their ancestors. Which axis is taken is not given by the rule when a filing cross-checks several: candidates are restricted to the rule's own axis list, and among those the system takes the one whose sum departs furthest from the reported value, discarding any that agrees with it. Inside that sum sits a heuristic of mine rather than of the rule: where the axis has more than two members and the definition linkbase declares none of them, so that no hierarchy is available, a member equal to the sum of the others is dropped as a subtotal. It is an arithmetic substitute for a hierarchy the filing does not state, and it is the ``subtotal fix'' row of Table~\ref{tab:dev}. Where the linkbase does declare the members and places them at one level, no member is dropped, because the rule then really does add the subtotal twice and that doubled sum is the finding. Behind the axis path a second-axis breakdown is tried as a last resort, which the rule's second assertion compares against a member's own fact rather than against the total. Nothing in the system reads a rule label. The paths are tried in a fixed order and the first that yields a value differing from the reported one wins, so a case where both an axis sum and a children sum disagree takes the axis sum by priority rather than by rule match. Section~\ref{sec:results} reports how often each fired.

Both rules are narrower than a bare reading of the linkbases would suggest, and my implementation follows them only part of the way. As published, DQC\_US\_0117 applies to statement networks defined with a hypercube and only over a listed set of axes; the system reproduces that axis list and restricts axis sums to the axes a statement role declares. DQC\_US\_0126 applies only to networks whose definition names a statement, considers only the current reporting period, and admits only facts carrying no dimensions; of these the system follows the statement-network restriction but implements neither the current-period restriction nor the no-dimensions restriction. So the system can flag a fact the rule never flagged, drawn from a period or a dimensional context the rule would not have examined. I did not measure how often this happens; Section~\ref{sec:gap} is why it is the first thing I would check next.

\subsection{Calculation}
A concept may carry calculation children in several extended-link roles. I rank the candidate roles by the number of children actually reported in the filing, breaking ties by the share of declared children that are present, and call the first of these the governing role. For a concept with calculation children I compute the signed sum of the children present for the asked period under that role, using the strict set of children declared in that role rather than a recursive expansion. Where the reported value and the computed value disagree, the computed value is the calculated answer.

Where no path produces a value differing from the reported one --- a concept with no children in the governing role and no axis to cross-check --- the chain ends in a sign flip and the system answers with the negation of the reported value. That is a guess and not a computation: nothing in the filing is consulted for it, and the organisers publish what the same strategy scores when it is applied to every case, 19.85\% on development~\cite{leaderboard}. It fires on one development case, which I cannot score because the development answers are not published. The branch should abstain, or fall back to the taxonomy balance as the organisers' rule-based baseline does.

\subsection{Candidate selection}
\label{sec:cand}
When several facts satisfy the asked concept and period, the system must choose one. On the practice set this choice is observable, because 21 of its queries repeat, 18 of them with different answers, each walking one list of findings. Ordering the candidates by the dimensional member name places the reference fact first in 14 of the 17 multi-candidate practice groups when the namespace prefix is kept and 16 of 17 when it is dropped, against 13 for the reported value and 10 for the fact identifier. The margin between the two member-name variants is two groups out of seventeen, which is inside the noise of a sample that size, and the two differ in output on exactly one case in each phase. The submitted system keeps the first candidate under the \emph{prefixed} ordering; the unprefixed variant was prepared but never scored. The ranked phases carry no repeated query, so a single fact from each such group reaches the set. Section~\ref{sec:limits} returns to the cases this does not settle.

\section{Results}
\label{sec:results}
\subsection{Development phase}
Table~\ref{tab:dev} reports the development scores. The system improves from 96.03\% to 96.62\% as the subtotal handling and the candidate ordering are corrected, each change confirmed against the judge rather than assumed. At 96.62\% the system labels 657 of 680 cases correctly, with 15 extraction errors and 8 calculation errors under the deterministic judge. The official judge scores the same submission at 96.32\% ACC, which is 655 of 680 and therefore 25 errors against the deterministic judge's 23. The entry stood 40th of 53 on the development leaderboard, where the leading entries scored between 98.5\% and 99.4\% and one reported 100\%. The development board is a live feed rather than a published file and I could not retrieve it while preparing this report. The development figures that come from the board rather than from a published page are my own record of 24~October 2026: the standings here, the probe readings of Section~\ref{sec:limits}, and the receipts of Table~\ref{tab:dev} that Section~\ref{sec:dev2test} compares against the test result. The official judge's error breakdown for this submission is not among them. The organisers' development baselines and Table~\ref{tab:judges} are published and dated where they are used, and the bound of Section~\ref{sec:bound} rests on the test board alone.

\subsection{Test phase}
\label{sec:test}
The published final standing~\cite{leaderboard} is 95.29\% accuracy, joint 15th of 23 teams, with SER 0.15\%, EER 1.03\% and CER 3.53\%. On 680 cases that is one structural error, seven extraction errors and twenty-four calculation errors, 648 correct.

The organisers record three submissions for this entry and score the last valid one. My own log covers two of them, neither of which is the configuration that scored 96.62\% on development, so I cannot confirm from my records which file was ranked. The comparison in Section~\ref{sec:dev2test} is written on the assumption that it was the development configuration, and should be read with that caveat.

\begin{table}[t]
\caption{Development phase, 680 cases, deterministic judge throughout. The first two rows are the organisers' published baselines~\cite{leaderboard}; the rest are this system's submission receipts. Three earlier receipts are omitted, two of them at 94.71\% and one at 95.00\%.}
\label{tab:dev}
\centering
\begin{tabular}{lccc}
\toprule
Configuration & ACC & EER & CER \\
\midrule
Rule-based baseline & 31.76\% & 10.88\% & 57.35\% \\
Sign-flip shortcut & 19.85\% & 10.88\% & 69.26\% \\
\midrule
This system, third distinct score & 96.03\% & 2.35\% & 1.62\% \\
\;+ subtotal fix & 96.47\% & 2.35\% & 1.18\% \\[2pt]
\;+ member-name ordering & \textbf{96.62\%} & 2.21\% & 1.18\% \\
\bottomrule
\end{tabular}
\end{table}

Each case is answered by one of three mechanisms. On the development set the calculated value is the signed sum of children in 461 cases, the sign flip of DQC\_US\_0015 in 161 cases, and the member sum of DQC\_US\_0117 in 57 cases; the remaining case falls through to a fallback.

\subsection{Development to test}
\label{sec:dev2test}
The official score fell from 96.32\% to 95.29\%, seven cases of 680. Two design choices were accepted because they improved the development leaderboard, with no held-out set to check them against, and a drop of this size is what fitting that feedback would produce. It is also about 1.4 standard errors at this accuracy and sample size, taking the single-sample standard error, so it is not distinguishable from sampling variation, and I report it as consistent with overfitting rather than as evidence of it. The two standings have different denominators, 53 entries on development and 23 teams on test, so the ranks are not comparable directly; as a share of the field, the entries ahead of this one fell from 39 of 52 to 14 of 22, from 75 to 64 per cent.

\subsection{The gap to the leading entries}
\label{sec:gap}
The published test board~\cite{leaderboard} makes the distance measurable, and it is a narrow board: 21 of the 23 ranked entries fall between 92.65\% and 97.79\%, a band of five accuracy points, with the remaining two at 79.12\% and 71.62\%. The top entries score 97.79\%, which is 665 of 680 against this system's 648, a gap of seventeen cases. Their breakdowns are of the same shape as this one but smaller: the leading entry records three extraction errors and twelve calculation errors, against seven and twenty-four here.

Decomposed by error class the gap is mostly calculation error, but that decomposition inherits the ambiguity of Section~\ref{sec:judges} and is not used here. The per-rule decomposition does not: the gap is four cases on DQC\_US\_0015, eight on DQC\_US\_0117 and five on DQC\_US\_0126. The leading entry loses fourteen of its fifteen cases on DQC\_US\_0015 and is perfect on the 464 DQC\_US\_0126 cases.

DQC\_US\_0015 carries the largest absolute number of errors for 18 of the 23 entries, this one included, and no entry exceeds 91.98\% on it. The gap sits in the two rules whose declared scoping this implementation does not reproduce (Section~\ref{sec:totals}): nine errors of 54 DQC\_US\_0117 cases against the leading entry's one, and five of 464 DQC\_US\_0126 cases against its none. Six entries on the board score below 100\% on DQC\_US\_0126, and this is one of them. That does not prove the omitted scoping is the cause, but it points at the part of the system I already know to be incomplete rather than at the part I spent the competition on.

By my record the development board's top entries sat between 98.5\% and 99.4\%, where the test maximum is 97.79\%.

\subsection{Neither the practice set nor the rule engine stands in for the ranked phases}
\label{sec:practice}
\label{sec:engine}
The practice set is the only split whose reference answers are in the participant's hands, so it is the only place where a score can be computed without submitting. Scored with the organisers' evaluator under the deterministic judge, the pipeline labels 193 of the 332 practice cases correctly, an accuracy of 58.13\%. The practice run on record predates the last two changes of Table~\ref{tab:dev} and so is not byte-identical to the ranked submission; my notes record that the candidate ordering does not move the practice score, but I did not re-score the final configuration here.

That score is not independent confirmation of the development one, and one episode shows why. If the benchmark's answers are the rules' own output, the open rule set should be able to adjudicate a disputed case, so I ran the live DQC engine, Arelle with the XBRL~US \texttt{xule} plug-in, over shipped documents on which it would run to completion. On a group of schedule-of-investments cases it returned no finding at all, which says the concept was not being flagged there and that the system was selecting a dimensional member where it should have taken the undimensioned total. Late in the development phase I made that change on the evidence of the practice set and of the engine together. It was consistent with the rule source, it agreed with the engine's verdict, and it did not lower the practice score. Submitted, it scored 96.47\% against 96.62\% for the unchanged system, with extraction errors rising from 15 to 16. At least one of the affected reference answers is a dimensional member after all. The change was reverted. The margin is a single case of 680, which is not in itself strong evidence; what it rules out is the hope that practice-set agreement plus the engine's verdict together make a submission unnecessary. The engine's verdict on the shipped filing and the benchmark's reference answer are not interchangeable.

A second and stronger reason not to read the practice score as a proxy is composition. On test the three rules appear as 162 DQC\_US\_0015, 54 DQC\_US\_0117 and 464 DQC\_US\_0126~\cite{leaderboard}; on practice they appear as 110, 120 and 102. DQC\_US\_0117 is 36\% of practice and 8\% of test; DQC\_US\_0126 is 31\% of practice and 68\% of test. The mechanism counts above show development follows the test composition, not the practice one. A practice score is a weighted average over a different mixture of rules, and the practice instances are truncated where the ranked ones are not. I treat agreement on the practice set as a weak signal.

\section{Analysis and Limits}
\label{sec:limits}
\subsection{What the residual error is}
The residual error concentrates in cases where a concept carries several facts for the asked period, each a legitimate flagged value under its own dimensional member, and the question names no member. The benchmark's construction keeps exactly one of them, and I could not determine from the shipped materials which one it keeps.

I tested the signals I was able to identify in the filing, and the list below is not exhaustive. The reference fact is not named in the question, not singled out by the supplied linkbases (all candidates appear in all of them), not always the first or the last in any document order, and not resolved by the rule's own exclusion lists beyond a handful of cases. On the practice set the submitted ordering recovers the reference fact in 14 of 17 groups, and the unsubmitted variant in 16, but seventeen groups is a small sample and the ranked phases are built from single draws. Two probes, submitted during the development phase and within its submission allowance, measured the ordering on the ranked data by deliberately spoiling extraction on a chosen set and reading the change off the leaderboard. Both probes predate the member-name ordering and measured the fact-identifier ordering that preceded it, which is the weakest of the four keys on practice at 10 of 17; the two orderings differ on 8 of the cases in question, so neither figure transfers to the submitted system unchanged. Because the scoring is hierarchical, each probe's count of correct cases is a lower bound on the number where the ordering recovered the reference fact.

The first probe covered a 15-case subset and found at least 5 right, against 5.2 expected if the choice within each group were uniform over its candidates: indistinguishable from chance. The second covered every development case the system marks as carrying more than one candidate, 31 of them, and returned at least 21 right. Rebuilding those 31 records gives candidate counts from 2 to 23, with no single-candidate case among them, and a uniform expectation of 10.7 with a standard deviation of 2.4; 21 is four standard deviations above that. So even the weakest key recovers the reference fact about twice as often as chance across ambiguous cases in general, while on the 15-case subset the same key does no better than chance. That subset is where the residual error sits, and the two figures are measured on differently selected sets, so they do not aggregate.

Two further checks bear on whether this is a property of the data rather than of my system. First, on the multi-candidate cases where the live DQC engine produced findings at all, it emits one for \emph{every} flagged candidate and singles out none; the rule itself does not prefer one. That is not universal: on the schedule-of-investments cases of Section~\ref{sec:engine}, which are also in this set, it produced no finding at all, which says those concepts were not being flagged rather than that the choice among them is undetermined. Second, I put a set of these cases to language models, with the complete case text and freedom to decide which fact the question meant. They were chosen for being ambiguous in the sense above, and they are ranked cases whose reference answers are not published, so the exercise yields no accuracy figure and I report none. What it did yield, in the one case I can characterise, is a model justifying its choice by the member's position in the orderings rather than by anything in the filing: given the whole document it reached for the same kind of heuristic the deterministic system uses, not for a signal the system was missing.

\subsection{The two judges classify the same errors differently}
\label{sec:judges}
The test breakdown does not say that arithmetic, rather than fact selection, is this system's problem, and the published rates do not support reading it that way. The organisers' baselines are the place to see why: one of them is published under both judges on the same 680 development cases~\cite{leaderboard}, which is the only such pair the organisers publish.

\begin{table}[t]
\caption{The organisers' rule-based baseline, same 680 development cases, scored by each judge. Counts derived from the published rates, read 19~November 2026: deterministic ACC 31.76\%, SER 0.00\%, EER 10.88\%, CER 57.35\%; official 33.09\%, 0.15\%, 3.97\%, 62.79\%~\cite{leaderboard}.}
\label{tab:judges}
\centering
\begin{tabular}{lcccc}
\toprule
Judge & A & S & E & C \\
\midrule
Deterministic & 216 & 0 & 74 & 390 \\
Official & 225 & 1 & 27 & 427 \\
\midrule
Difference & $+9$ & $+1$ & $-47$ & $+37$ \\
\bottomrule
\end{tabular}
\end{table}

Table~\ref{tab:judges} reports 47 fewer extraction errors and 37 more calculation errors under the official judge, with 9 more correct and one more structural. These are net column differences and not a case-level flow, so what they give is a range rather than a figure: at least 46 reported values that the deterministic judge rejected were accepted, and at most 74, since the official judge books only 27 extraction errors in total and the deterministic judge only 74. The extra structural error is a reminder that the official judge is not uniformly the more permissive of the two.

One part of this follows from the label definitions alone. The scheme is hierarchical: a case whose reported value is rejected is never examined for its calculated value. So for any judge, accepting a reported value that is in fact wrong moves a still-failing case out of the extraction column and into the calculation column. An official calculation-error rate therefore cannot be decomposed into arithmetic failures and fact-selection failures by the reader, whatever the two judges happen to do. The test breakdown of Section~\ref{sec:test} cannot be used to argue that fact selection is a small problem and arithmetic a large one, and the fact-selection failures of Section~\ref{sec:limits} may be part of the twenty-four calculation errors rather than a separate and smaller class.

The organisers read the same 47 cases differently: the official judge ``accepts 7 points more of the same extractions'' because ``it reads values that mean the same number but are written differently''~\cite{leaderboard}, which makes the shift a property of the baseline's output format rather than of either judge's standard. The two accounts are not distinguishable from the published rates, and they differ in what generalises, since on the organisers' reading a submission that emits canonical numeric values should show no such shift at all. I cannot settle it without the per-case labels.

A second pair falsifies the general form of the claim. The organisers state that ``a deterministic figure is a lower bound on the official one''~\cite{leaderboard}. On this system's development submission it is not: the deterministic judge scores 96.62\% and the official judge 96.32\%, two cases lower.

\subsection{What the per-rule accuracies do settle}
\label{sec:bound}
The \emph{class} of the remainder cannot be read off the published rates, for the reason just given. Its \emph{cost} can. Accuracy is untouched by the relabelling, and the organisers publish per-rule accuracy for every ranked entry~\cite{leaderboard}. This system's 32 test errors fall as 18 on DQC\_US\_0015 (162 cases), 9 on DQC\_US\_0117 (54) and 5 on DQC\_US\_0126 (464).

The first figure settles something the error-class rates cannot, on one premise that has to be stated plainly because everything after it depends on the premise holding. On a DQC\_US\_0015 case the calculated answer is the negation of the reported one, so once the fact is chosen no arithmetic is left to get wrong. The practice split is the only split whose answers are published, and there the identity is exact and specific to this rule: the reference calculated value is the negation of the reference reported value in 110 of the 110 DQC\_US\_0015 cases, against 1 of 120 for DQC\_US\_0117 and 6 of 102 for DQC\_US\_0126. It is also what the rule is for, since \emph{Negative Values} flags a sign and the correction is that sign, and the organisers note in passing that a third of the practice set is ``the one rule whose answer is always the negation''~\cite{leaderboard}. The ranked answers are not published, so carrying the identity to the 162 DQC\_US\_0015 cases of the test split is an inference from the rule's definition and from 110 of 110 on practice, not a published fact. It is the one assumption this section rests on.

The counts behind this are reconstructed from published percentages, and they check themselves. The per-rule accuracies give 18, 9 and 5 errors on the three rules; the class rates give 1 structural, 7 extraction and 24 calculation. Both sum to 32, and the rule case counts sum to 680. A rounding artefact is ruled out twice over. One case is 0.617 of a percentage point over 162, 1.852 over 54 and 0.147 over 680, so for every published figure on the board exactly one integer rounds to it; and independently of that, the two sums agree in all 23 rows.

Each case carries exactly one label, and only eight structural or extraction labels exist in total, so at most eight of the 18 DQC\_US\_0015 errors can be one of those. \emph{At least ten of this entry's twenty-four calculation errors therefore sit on cases whose calculated value is determined by the reported one.} The calculated value there carries no independent arithmetic content: either a wrong reported value was accepted and the failure moved into the calculation column, or the case was routed down the children-sum path instead of the sign path. Arithmetic was performed in the second of these, but not the arithmetic the question asked for.

The same subtraction applies to every row of the board, and at the top it bites harder than here (Table~\ref{tab:bound}). The two joint-first entries make 14 errors on DQC\_US\_0015 against three and two structural-or-extraction labels in total, so at least 11 of one entry's 12 calculation errors and 12 of the other's 13 sit on determined cases. About 92 per cent of the calculation error of the best systems on this task is not arithmetic. The bound is positive for 22 of the 23 ranked entries and covers at least half the calculation error for nine of them. No entry escapes it by being accurate on that rule: none scores above 91.98\% on DQC\_US\_0015 and every one makes at least 13 errors there, which is also why that rule carries the largest absolute error count for 18 of the 23 entries.

None of this depends on which of this entry's three submissions was ranked, on Table~\ref{tab:judges}, or on any assumption about what the official judge does beyond the quoted definition. It is subtraction on the published board, and it is the account of the calculation-error count that the two judges leave open.

\begin{table}[t]
\caption{Calculation errors that cannot be arithmetic. Errors per rule and per class are reconstructed from the published accuracies and rates~\cite{leaderboard}; $e_{15}$ is the error count on DQC\_US\_0015, $S{+}E$ the structural and extraction labels available in total, and the bound is $e_{15}-(S{+}E)$, the least number of calculation errors that must fall on cases whose calculated value is determined by the reported one.}
\label{tab:bound}
\centering
\begin{tabular}{@{}lccccc@{}}
\toprule
Entry & ACC & $e_{15}$ & $S{+}E$ & $C$ & bound \\
\midrule
Polymanes (joint 1st)  & 97.79\% & 14 & 3  & 12 & $\geq 11$ \\
FinVeritas (joint 1st) & 97.79\% & 14 & 2  & 13 & $\geq 12$ \\
xai (3rd)              & 97.50\% & 16 & 4  & 13 & $\geq 12$ \\
This system (joint 15th) & 95.29\% & 18 & 8  & 24 & $\geq 10$ \\
\bottomrule
\end{tabular}
\end{table}

I found no signal in the shipped materials that selects the reference fact, and neither the engine nor the models offered one on the cases I put to them. That does not establish that no such signal exists: the samples are small, they were chosen for being ambiguous rather than at random, and the leading entries score above this one.

\section{Conclusion}
Modelling the three DQC rules directly gives a compact, interpretable system with no learned parameters, which reaches 96.62\% on the development set under the offline judge and 95.29\% on the test set under the official one. Its residual error concentrates in cases where several facts are equally flagged for the asked period and nothing in the shipped materials settles which one the reference keeps. For anyone entering a task of this shape, the one lesson that transfers is procedural, and it is uncomfortable. The practice split's rule composition is 28 points away from the ranked split on one rule and 37 on another, so practice agreement does not carry; and the ranked answers are not released, so a held-out set cannot be built either. Every check of a design choice costs a submission from the daily allowance. That makes the allowance the scarce resource of the competition and worth budgeting before the work starts, rather than spent as this entry spent it, on changes already believed to be right.

The second finding concerns the evaluation. Because the scheme is hierarchical, accepting a reported value that is in fact wrong moves a still-failing case out of the extraction column and into the calculation column. A participant reading the official rates therefore cannot tell how much of the calculation error is arithmetic and how much is a wrongly chosen fact. The organisers' baseline, scored by both judges on the same 680 cases, shows 47 fewer extraction errors and 37 more calculation errors under the official one, which is consistent with that mechanism without establishing it, and the organisers themselves read those cases differently. What is not ambiguous is the per-rule accuracy, which the relabelling does not touch, and it bounds the effect from the published numbers alone. Of this entry's 32 test errors 18 fall on DQC\_US\_0015, whose answer is the negation of the reported value in all 110 published practice cases and, by the inference set out above, on the ranked split too, while seven extraction labels and one structural label exist in total, so at least ten of its 24 calculation errors sit on cases where no arithmetic was left to get wrong. The bound is positive for 22 of the 23 ranked entries, and it is tightest where the systems are best: at least 11 of 12 and 12 of 13 calculation errors for the two joint-first entries. At the top of the board, the calculation-error rate is almost entirely not measuring what its name says.

Two suggestions follow for the benchmark. The first is about the metric rather than about how much of it is published: a calculation-error rate measured hierarchically cannot report what its name says, because every accepted-but-wrong reported value lands in it. Scoring the calculated value over the subset where the reported value matches the reference, alongside the hierarchical figure, would separate the two for every entry at no extra annotation cost. Per-rule accuracy should keep being published, since the relabelling does not touch it, and it is what carried the argument above. And the selection step is avoidable: carrying the dimensional member into the question, or drawing complete finding groups rather than single survivors, would remove it. On the development set it affects 31 cases of 680, so it is an avoidable source of ambiguity in the task's construction rather than a limit on the task itself.

\section*{Reproducibility}
The system is Python~3.10 and imports nothing outside the standard library. Parsing and record construction over the 680 development cases take 1.4\,s single-threaded with a peak resident set of 113\,MB, and the policy step a further 0.03\,s; there is no model, no training step and no external service. Re-running the pipeline described here against the published development inputs reproduces the submitted file exactly, all 680 rows and the same mechanism counts. It is not reproducible by a third party, because the source code is not released: it carries exclusion lists and orderings I intend to keep working on, and I would rather release it once than release it early.

The DQC engine check used Arelle with the XBRL~US \texttt{xule} plug-in version 3.0.30055, against the rule sources on the \texttt{dqc\_us\_rules} master branch as read in October 2026. The exclusion lists this system reproduces change between rule-set releases, so a later reader should pin a release rather than take the latest.

The model probe is recorded in my notes rather than in the run log that covers the rest of this work; the limits of the other two checks are stated where each is used.

\end{document}